\documentclass[conference]{IEEEtran}

\usepackage{graphicx}       

\usepackage{algorithm}
\usepackage{algpseudocode}

\usepackage{amssymb}

\usepackage{booktabs}

\usepackage{extarrows}

\title{Rate-Distortion Guided Knowledge Graph Construction from Lecture Notes Using Gromov-Wasserstein Optimal Transport
}

\author{
\centering
\IEEEauthorblockN{Yuan An, Ruhma Hashmi,  Michelle Rogers, Jane Greenberg}
\IEEEauthorblockA{\textit{College of Computing and Informatics} \\
\textit{Drexel University}\\
Philadelphia, PA 19104, USA \\
\{ya45, rh927, mlr92, jg3243\}@drexel.edu}
\\[1em] 
\IEEEauthorblockN{Brian K Smith}
\IEEEauthorblockA{\centerline{\textit{Lynch School of Education and Human Development \&
Department of Computer Science}} \\
\textit{Boston College}\\
Boston, MA 02467, USA\\
b.smith@bc.edu}
}

\begin{document}
\maketitle


\begin{abstract}
Task-oriented knowledge graphs (KGs) enable AI-powered learning assistant systems to automatically generate 
high-quality multiple-choice questions (MCQs). Yet converting unstructured educational materials, 
such as lecture notes and slides, into KGs that capture key pedagogical content remains difficult. 
We propose a framework 
for knowledge graph construction and refinement grounded in rate-distortion (RD) theory and optimal 
transport geometry. In the framework, lecture content is modeled as a metric-measure space, capturing semantic 
and relational structure, while candidate KGs are aligned using Fused Gromov-Wasserstein (FGW) 
couplings to quantify semantic distortion. 
The rate term, expressed via the size of KG, 
reflects complexity and compactness. Refinement operators (add, merge, 
split, remove, rewire) minimize the rate-distortion Lagrangian, yielding compact, information-preserving 
KGs. Our prototype applied to data science lectures yields interpretable RD curves and shows that 
MCQs generated from refined KGs consistently surpass those from raw notes on fifteen 
quality criteria. This study establishes a principled foundation for information-theoretic 
KG optimization in personalized and AI-assisted education.
\end{abstract}

%
%
\section{Introduction}
\label{sec:introduction}

An AI-powered learning assistant (AILA) system \cite{an2024lecture}
can automatically generate educational questions such as 
multiple choice questions (MCQs) from teaching materials including lecture notes and slides
\cite{an2025enhancing,automated-edu-question-generation}. Despite the convenience and time-saving benefits,
LLM-generated MCQs present critical quality issues, including hallucinations, factual mistakes, 
trivial questions, ambiguous options, and easy distractors \cite{an2025enhancing}. Recent research demonstrates that integrating 
knowledge graphs (KGs) with large language models (LLMs) significantly improves the factual 
accuracy, explainability, and reasoning capabilities of LLM-powered answers 
\cite{wang-etal-2024-infuserki,xu-harnessing-2025}. 
We propose that integrating knowledge graphs (KGs) into AI-powered learning assistant (AILA) 
systems will also enhance the generation of multiple-choice questions (MCQs).  
Furthermore, we aim to construct task-oriented KGs that employ a set of pedagogically 
grounded relationships, such as \texttt{prerequisite-of}, \texttt{contrastingWith}, and \texttt{example-of}, 
specifically designed for educational applications. These relations enable curriculum-aware 
question generation that is both cognitively engaging and intellectually demanding.

\begin{figure}[!t] 
        \centering 
        \includegraphics[width=\linewidth]{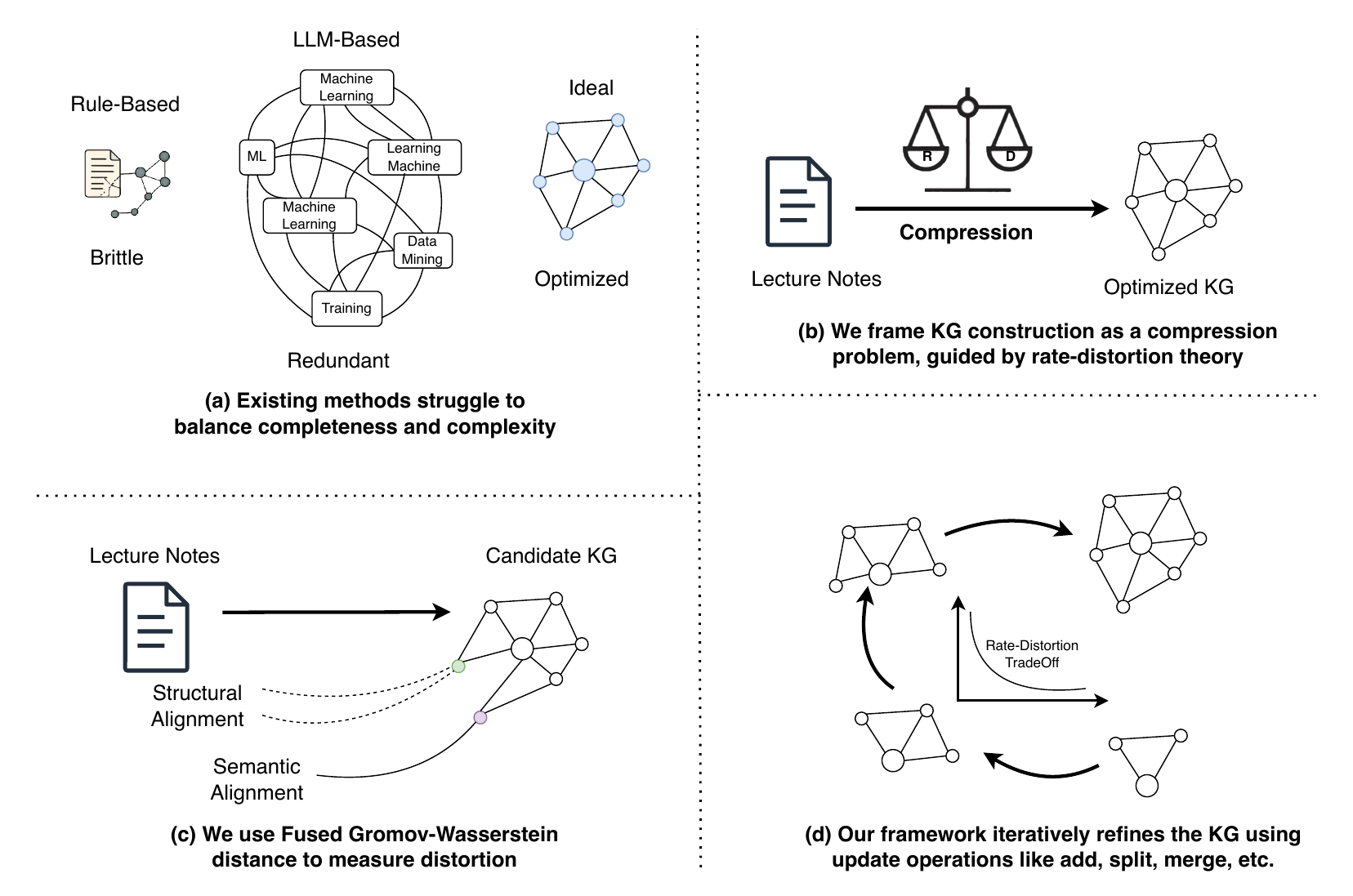} 
        \caption{A Rate–Distortion Framework for Knowledge Graph Construction} 
        \label{fig:overview} 
\end{figure}

Extracting KGs from unstructured lecture notes and slides remains a challenging problem.
Automatic approaches exist, yet they struggle with balancing completeness and 
complexity (Figure \ref{fig:overview} (a)). Rule-based pipelines \cite{Schneider-decode-KG-2022}
(e.g. using grammar patterns and 
heuristics) can extract concepts and relations, but they tend to be brittle (hard to generalize beyond their designed rules). 
Machine learning and more recent large language model (LLM) based methods
\cite{zhong2023comprehensive} 
can extract concepts and relations from text, but they frequently yield incomplete KGs with missing
elements or bloated ones  with redundant nodes 
and facts, as well as inconsistencies or errors in hierarchy. 
Their performance can vary across domains, often introducing 
spurious relations or semantic drift in the generated knowledge graphs
\cite{liu-etal-2024-shot,saeedizade2024navigating}. In short, existing approaches either over-generate content 
or require heavy curation to ensure correctness. 
There is currently no principled mechanism to balance complexity 
against how well the knowledge graph captures the source content for a certain task.

In this paper, we seek a systematic way to manage 
the trade-off between a knowledge graph’s size and its 
accuracy in representing the lecture content. 
To address the problem, we introduce a novel framework grounded in rate-distortion (RD) theory \cite{harell2025rate} and 
optimal transport \cite{COT-DS} to guide the construction of knowledge graphs (Figure \ref{fig:overview} (b)). 
In information theory, rate-distortion theory 
provides a formal way to quantify the minimal distortion (information loss) achievable for a given 
compression rate (complexity) \cite{harell2025rate}. 
We leverage this idea by treating the lecture content as a ``source” and the 
knowledge graph as a compressed ``representation,” 
and then using optimal transport \cite{COT-DS} to align them in a way that quantifies the distortion. 
In particular, we adopt the fused Gromov–Wasserstein (FGW) distance
\cite{vayer2020fgw}  
to measure how far a candidate knowledge graph is from the lecture content in 
terms of both semantic content and relational structure (Figure \ref{fig:overview} (c)). 
By iteratively refining the knowledge graph through local operations,
we can obtain a knowledge graph that 
faithfully represents the lecture material (Figure \ref{fig:overview} (d)).

Our work makes the following contributions:
\begin{itemize}
	\item We present a rigorous formulation where lecture notes (source) and an knowledge 
	graph (representation) are modeled as metric spaces. The knowledge graph construction problem 
	is cast as finding a representation that balances information loss for a given complexity budget. 
	We define rate $R$ as the size of the knowledge graph  
	and distortion $D$ as a distance measuring how much the knowledge graph 
	diverge from the source content. This framework lets us explicitly balance compression 
	vs. fidelity by optimizing a Lagrangian objective: $L = R + \beta D$.

	\item We adopt a fused Gromov-Wasserstein (FGW) optimal transport approach to evaluate distortion. 
	The FGW distance extends traditional Wasserstein distance \cite{Villani2009} to simultaneously account for both 
	structural and semantic differences. To optimize the objective, we analyze the coupling matrix associated with 
	FGW distance for iterative KG updates.  
	
	\item We propose a set of knowledge graph update operations: adding a new concept, 
	merging similar concepts, splitting a coarse 
	concept, pruning an unneeded concept, and rewiring relationships. Each operation evaluated by its impact on the 
	$L = R + \beta D$ objective. By applying these operators iteratively and accepting changes that improve the trade-off, the knowledge 
	graph is refined toward an optimal rate-distortion balance. 
	
	\item We develop a prototype system and demonstrate it on a set of data science lecture notes. Starting from a rudimentary 
	knowledge graph automatically bootstrapped from lecture notes, our RD-guided process produces gradually more 
	representative knowledge graph that attempts to cover key concepts. We report quantitative metrics (coverage of content and 
	quality of MCQs) and show the evolution of the rate-distortion curve during 
	refinement, illustrating how an optimal “knee point” is reached. This case study highlights the potential of our 
	approach for improving educational knowledge engineering.
\end{itemize}

The rest of the paper is organized as follows. 
Section \ref{sec:related_work} discusses related work.
Section \ref{sec:theoretical_framework} develops theoretical frameworks.
Section \ref{sec:implementation} presents the implementation of the prototype.
Section \ref{sec:experiments} describes the experiments and the resulting findings.
Finally, Section \ref{sec:conclusion} concludes with a summary of the 
contribution to the improvement of AI powered learning systems
and student learning outcomes.

%
%
\section{Related Work}
\label{sec:related_work}

\subsection{Knowledge Graph Extraction from Text}

Automatically extracting knowledge graphs relies on 
named entity recognition (NER) \cite{Seow2025review} 
and relation extraction (RE) \cite{zhao2024comprehensive}.
Traditional approaches include using 
linguistic heuristics \cite{norabid2022rule}, 
and more recently using deep learning or LLM 
\cite{Zhao2023relation,hu-etal-2025-large,al2020automatic,LIANG2024870,bian2025llmpowered}. 
By leveraging vast semantic knowledge, LLMs can be prompted to generate lists of key topics or to predict relationships
between terms in a corpus. 
However, studies have also noted 
that LLM-based knowledge graph generation can produce inconsistent or logically 
invalid hierarchies \cite{FathallahSA24,BakkerSB24}. 
Recent research on LLM-assisted knowledge graph generation 
highlights that while these models can accelerate concept extraction, the quality control 
often requires additional algorithms or expert review \cite{Ghanem2025}.

\subsection{Information-Theoretic Approaches in Information Representation}

\begin{figure}[h] 
        \centering 
        \includegraphics[width=\linewidth]{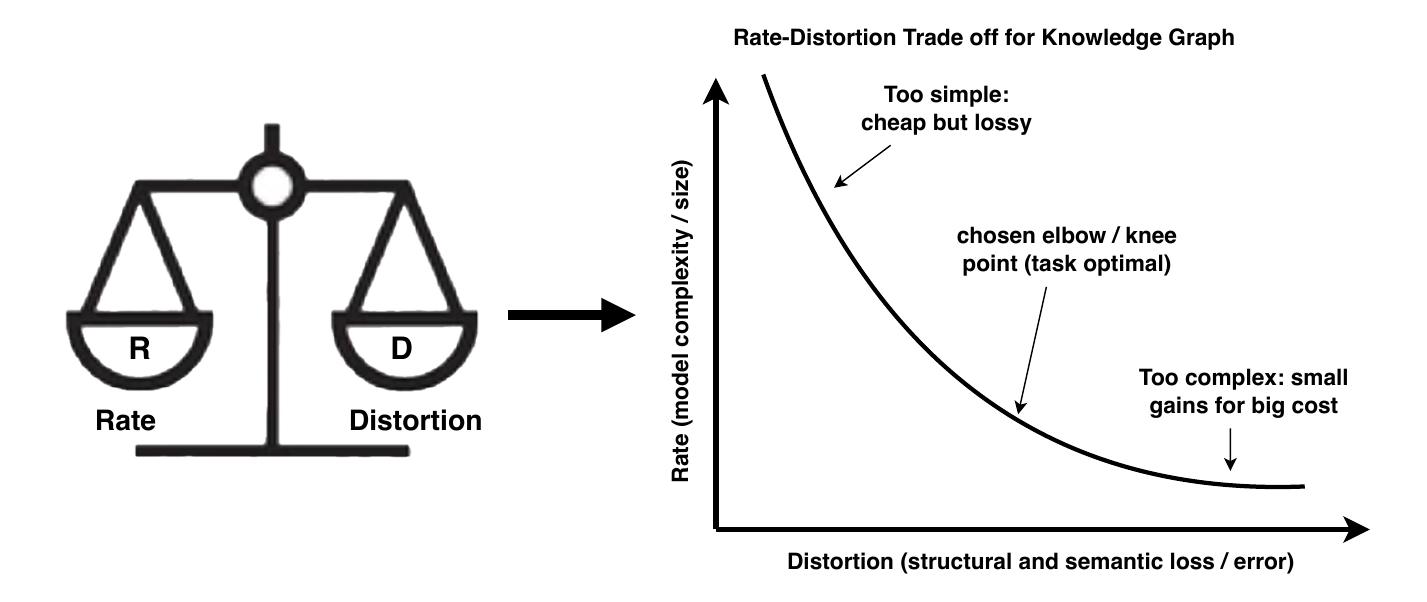} 
        \caption{Balance between Rate and Distortion} 
        \label{fig:rd_balance} 
\end{figure}

Our methodology is inspired by information-theoretic techniques, especially rate-distortion theory, which is a cornerstone 
of lossy compression. Rate-distortion theory formalizes the trade-off between the compression rate (how succinctly data 
is encoded) and the distortion (error or information loss from compression) \cite{niu2025rateDistortion} (Figure \ref{fig:rd_balance}). 
The classic result is the rate-distortion 
curve, which gives the minimum achievable distortion for any given rate \cite{Martínez_2019}. 
Bardera et al. \cite{bardera2017rate} proposed an information-theoretic 
clustering framework based on rate-distortion, achieving a maximally compressed grouping of data with minimal signal distortion. 
Our work is novel in applying an RD perspective to 
knowledge graph engineering. We treat the knowledge graph as a compressed knowledge representation of the source. 
This formalism provides a clear target, the RD optimum, which we hypothesize will correspond to an knowledge 
graph that is neither too large (including every minor detail) nor too small (missing important concepts). By plotting an 
RD curve as we refine the knowledge graph, we can identify the “knee” (elbow point) where adding more concepts yields 
diminishing reduction in distortion (see Figure \ref{fig:rd_balance}). This yields a principled stopping criterion for knowledge graph complexity, rather 
than relying on ad-hoc thresholds.

\subsection{Optimal Transport for Structured Data}

Optimal transport (OT) \cite{COT-DS} provides a way to measure distances between probability distributions by finding an optimal 
coupling (mapping) that minimizes the cost of transporting mass from one distribution to another. 
A direct distance called Wasserstein distance \cite{Villani2009} assumes both distributions live in the same metric space. 
In our scenario, lecture content and knowledge 
graph elements are not in the same space, one is essentially text/knowledge in lecture order, the other is a graph of 
concepts. This calls for Gromov-Wasserstein (GW) distance, which is a form of OT designed to compare two metric 
spaces even if they are different. 
Furthermore, we use the fused Gromov–Wasserstein (FGW) distance \cite{vayer2020fused}, 
which extends GW by also considering feature/attribute 
similarity in addition to relational structure. 
In our case, lecture notes have textual 
content and the knowledge graph has concepts that can be represented by definitions or embeddings. FGW allows us to align these 
two both by semantic content and by structural position. 

\subsection{Knowledge Graphs and AI in Education}

Knowledge graphs and concept maps play an important role in many educational applications 
\cite{AbuSalih2024,aytekin2024ace}. An knowledge graph 
of a domain can serve as the domain model for an intelligent tutoring system \cite{canal2024educational}, encoding the 
key concepts a student must learn and the relationships  that structure the curriculum. 
Recently, LLMs play an important role in classrooms \cite{witsken2025llms}.
Studies show that LLMs enhance student learning and performance\cite{yusof2025chatgpt}
and excel at generating diverse types of questions, reducing the workload of instructors \cite{hu2025exploring}.
However, the direct use of LLMs faces major challenges with output quality, pedagogy, and reliance 
on human oversight. AI-generated MCQs require extensive validation due to ambiguity or poor alignment 
\cite{yusof2025chatgpt,witsken2025llms}.

%
%
\section{Theoretical Framework}
\label{sec:theoretical_framework}

\subsection{Rate-Distortion Formalization of Knowledge Graph Construction}

We formalize the lecture notes and the knowledge graph as two metric measure spaces and define a rate-distortion 
objective between them. Let the source be the lecture content, denoted $S = (Z, d_Z, \mu_Z)$, 
where $Z$ is the set of elements (e.g. segments of text, topics, or concept mentions in the lectures), 
$d_Z$ is a distance metric capturing the similarity between lecture elements (this could be based on semantic 
similarity or the structure of the lecture sequence), 
and $\mu_Z$ is a probability measure over $Z$ (for instance, weighting segments by their length 
or importance). Likewise, the knowledge graph is the reproduction (compressed representation), 
denoted $O = (V, d_V, \mu_V)$, with $V$ the set of concept nodes, $d_V$ a distance metric on 
knowledge graph concepts (e.g. shortest-path distance in the concept graph or a semantic distance if available), 
and $\mu_V$ a measure over $V$ (perhaps weighting each concept by its relevance or simply a uniform 
measure over knowledge graph nodes).

Intuitively, $S$ contains all the information in the lectures, and $O$ is our attempt to encode that information in 
a knowledge graph. In classical rate-distortion theory, one seeks to maximize the mutual information between 
source and code under a distortion constraint, or vice versa minimize distortion for a given information rate. 
Here, the rate $R$ is defined as the mutual information $I(Z;V)$ between the source and code. This can be thought of as how 
much information about the lecture content is retained in the knowledge graph. Consequently, $R$ will increase as the 
knowledge graph becomes more complex (more nodes and links can capture more distinctions from $Z$). 
In our implementation, we approximate $R$ using the description length of the knowledge graph, 
such as its total number of nodes and edges.
Conceptually, a larger, more detailed graph requires more bits to describe, and therefore corresponds to 
a higher information rate.

\begin{figure}[h] 
        \centering 
        \includegraphics[width=\columnwidth]{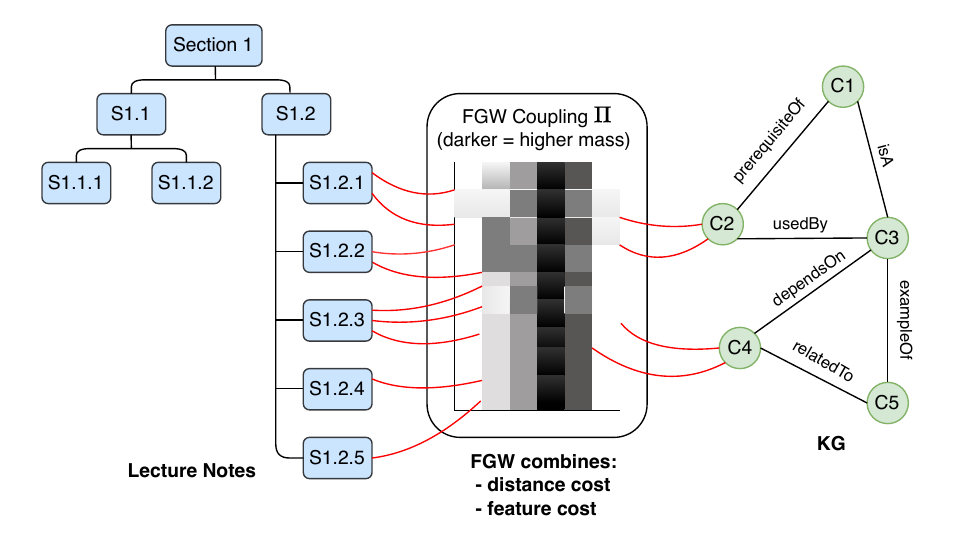} 
        \caption{Mapping the lecture notes on left-hand side to the knowledge graph (KG) on right-hand side 
        via FGW coupling $\Pi$ in the middle. The coupling matrix $\Pi$ is computed 
        through minimizing both structural and feature mismatches.} 
        \label{fig:coupling} 
\end{figure}

\begin{figure*}[!t]
\begin{equation}
\label{eq:distortion}
D(S,O) \;=\; 
\min_{\pi \in \Pi(\mu_Z, \mu_V)} \left[
(1-\lambda) \sum_{z,z' \in Z} \sum_{v,w \in V} 
\big| d_Z(z,z') - d_V(v,w) \big|^2 \,\pi(z,v)\,\pi(z',w)
+ \lambda \sum_{z \in Z} \sum_{v \in V} c_{\text{feat}}(z,v)\,\pi(z,v)
\right]
\end{equation}
\end{figure*}

The distortion $D$ between the lecture space and knowledge graph space is defined using a coupling 
$\pi$ between $Z$ and $V$. Formally, let $\Pi(\mu_Z, \mu_V)$ be the set of all joint distributions (couplings) 
with marginals $\mu_Z$ on $Z$ and $\mu_V$ on $V$. Each $\pi(z, v)$ in the coupling represents some 
amount of “mass” matching lecture element $z$ to knowledge graph concept $v$ (see Figure \ref{fig:coupling}). 
The distortion for a given coupling is defined as a combination of structural mismatch and feature mismatch,
which is computed as fused Gromov Wasserstein (FGW) distance through optimal transport.

\textbf{Structural distortion:} For any two lecture elements $z, z' \in Z$ that are mapped to concepts 
$v, w \in V$ via $\pi$, we incur a cost for how different the distances are: $|d_Z(z, z') - d_V(v, w)|^2$. This term 
is summed over all pairs, weighted by $\pi(z,v)\pi(z',w)$. If the lecture had two topics that were 
very far apart (unrelated) but the knowledge graph maps them to two concepts that are close 
(say, linked as prerequisites), that increases distortion, and vice versa.

\textbf{Feature/semantic distortion:} For each pairing of a lecture element $z$ to a concept $v$, we also consider a 
feature difference cost $c_{\text{feat}}(z,v)$. This could be defined, for example, 
as $1 - \cos(\text{embedding}(z), \text{embedding}(v))$ if we embed lecture text and concept labels in a 
vector space. If a lecture segment’s content is very different from the concept to which the segment is aligned, that incurs distortion. 
We weight this by $\pi(z,v)$ as well, and sum over all $z,v$.

A coupling $\pi$ that perfectly aligns the lecture to the knowledge graph would make both terms small: it would pair 
up lecture segments and concepts in a way that preserves distances (structure) and matches content (features). 
Of course, a perfect alignment may not exist if the knowledge graph is too simple. 
We therefore define the distortion $D(S,O)$, i.e., fused Gromov Wasserstein (FGW) 
distance, as the minimum distortion over all possible couplings $\pi$ in 
Equation~(\ref{eq:distortion}).

\textbf{Explanation of Equation~(\ref{eq:distortion})}: The hyperparameter $\lambda$ is a weight to 
balance the structural mismatch term vs. the feature mismatch term.
The first sum computes the GW discrepancy between $(Z,d_Z)$ and $(V,d_V)$, 
while the second term adds the feature-based cost (hence ``fused"). Finding the minimizing $\pi$ is an optimal 
transport problem. It finds which lecture pieces should align with which concepts to best preserve relative 
distances and semantic content. The resulting $D$ increases if the knowledge graph is missing concepts 
(so some lecture content has to be matched suboptimally) or if the knowledge graph’s structure is very unlike 
the lecture’s structure.

\textbf{Optimization Objective}: 
With $R$ and $D$ defined, we combine them into a single rate-distortion objective:
\[
L = R + \beta D
\]

where $\beta$ is a Lagrange multiplier (or trade-off parameter) that controls the relative importance of 
distortion vs. rate. By adjusting $\beta$, we can emphasize fidelity (low distortion) or parsimony 
(low complexity). Tracing $L$ as $\beta$ varies would theoretically produce the rate-distortion curve 
for this source. In practice, we might pick a particular $\beta$ that reflects how much loss we are 
willing to tolerate for reducing complexity, or we might seek the “knee point” where $D$ starts 
rising sharply if $R$ is decreased further.

\textbf{Interpretation of the Objective:} At one extreme, $\beta \to \infty$ would force distortion to zero, which would 
likely require $R$ to be very high (meaning the knowledge graph memorizes the lectures with many 
concepts and links, essentially one concept per lecture fragment). At the other extreme, $\beta \to 0$ would 
ignore distortion and minimize $R$, which is achieved by a trivial knowledge graph (maybe one node 
representing everything), obviously with high information loss. For a moderate $\beta$, minimizing 
$L$ yields a knowledge graph that compresses the lecture content as much as possible while keeping 
distortion within acceptable bounds. This formalism gives us a target for optimization and a way 
to quantitatively evaluate any candidate knowledge graph.

\subsection{Knowledge Graph Refinement via RD Optimization}

How do we search the best knowledge graph to represent the lecture notes? 
The search space of all possible 
knowledge graphs is enormous (combinatorial in concepts and relations). Our strategy is to start 
with an initial knowledge graph and then iteratively refine it, using local operations that change the 
knowledge graph.

\begin{figure}[h] 
        \centering 
        \includegraphics[width=\columnwidth]{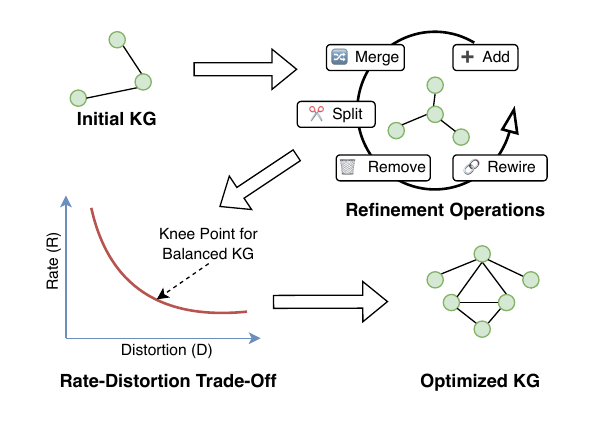} 
        \caption{Iterative Rate-Distortion guided refinement of a Knowledge Graph representing lecture notes. 
        The process begins with an initial graph and applies local edit operations 
        (add, merge, split, remove, rewire) to minimize combined objective. 
	The resulting graph achieves an optimal tradeoff between rate (R) and 
	distortion (D)} 
        \label{fig:operations} 
\end{figure}

We can think of $R$ roughly as increasing with each additional concept or relationship (complexity) 
and $D$ decreasing as the knowledge graph becomes richer (since it can explain more of the lecture). 
The $L$ combines them, so an operation is beneficial if it sufficiently lowers $D$ to justify any increase 
in $R$. Concretely, we implement the following refinement operators, inspired by typical 
knowledge graph editing actions (see Figure \ref{fig:operations}):
\begin{itemize}

	\item \textbf{Add a concept node:} If the current knowledge graph fails to cover a significant portion of the 
	lecture content (e.g., an entire subtopic or cluster of material has high distortion), we consider adding a 
	new concept to represent that missing area.

	\item \textbf{Merge nodes:} If two concept nodes in the knowledge graph are largely redundant or always mapped 
	to the same set of lecture items, merging them into one can reduce complexity without much increase in distortion. 

	\item \textbf{Split a node:} The inverse of merge. If an knowledge graph node is trying to cover content that is too heterogeneous 
	(for instance, one concept node is aligned to two very different topics in the lectures), then the distortion term will be high for that 
	node’s alignment. Splitting it into two more specific concepts can lower distortion.
		
	\item \textbf{Prune (remove) a node:} If a concept node has very low probability mass $\mu_V$ or very few lecture elements 
	aligning to it, it might be unnecessary. Perhaps it was a concept that doesn’t actually appear much in the lectures or a very 
	fine detail. Removing it can save complexity ($R$ down) while only slightly increasing distortion.
	
	\item \textbf{Rewire relations:} This operation adjusts the edges/links in the knowledge graph without changing 
	the set of nodes. If the alignment suggests that the current prerequisite structure is suboptimal (e.g., lecture content 
	shows concept A is always introduced before B, but the knowledge graph had them unrelated), we might add a prerequisite 
	link $A \to B$. Conversely, if two concepts are linked as if related but the lecture content treats them independently (large $d_Z$), 
	perhaps that link should be removed or changed.

\end{itemize}

\subsection{Evaluation Metrics for the Knowledge Graph}
\label{subsec:evaluation_metrics}

Theoretically, we define several metrics for assessing the quality of the knowledge graph.

\begin{itemize}
	\item \textbf{Coverage:} This measures what fraction of the lecture content is ``explained” or aligned 
	by the knowledge graph given a certain level of feature mismatch. 
	
	\item \textbf{Pedagogical flow alignment:} This compares the structure of the knowledge graph to the sequence of 
	how content is taught. We can quantify this by looking at the ordering of 
	concepts in the lecture timeline vs. the knowledge graph graph structure. 

	\item \textbf{Retrieve practice quality:} This tests how the knowledge graph helps generate high-quality retrieve practice questions such as 
	MCQ for learning assistance. 

	\item \textbf{RD curve:} This measures how refinements change KG to find a balanced rate and distortion point.
	
\end{itemize}

%
%
\section{Implementation}
\label{sec:implementation}

Our implementation begins with a cold-start phase that converts Markdown 
structured lecture notes into a metric–measure space and  bootstraps an initial knowledge graph.

\begin{figure}[h] 
        \centering 
        \includegraphics[width=\columnwidth]{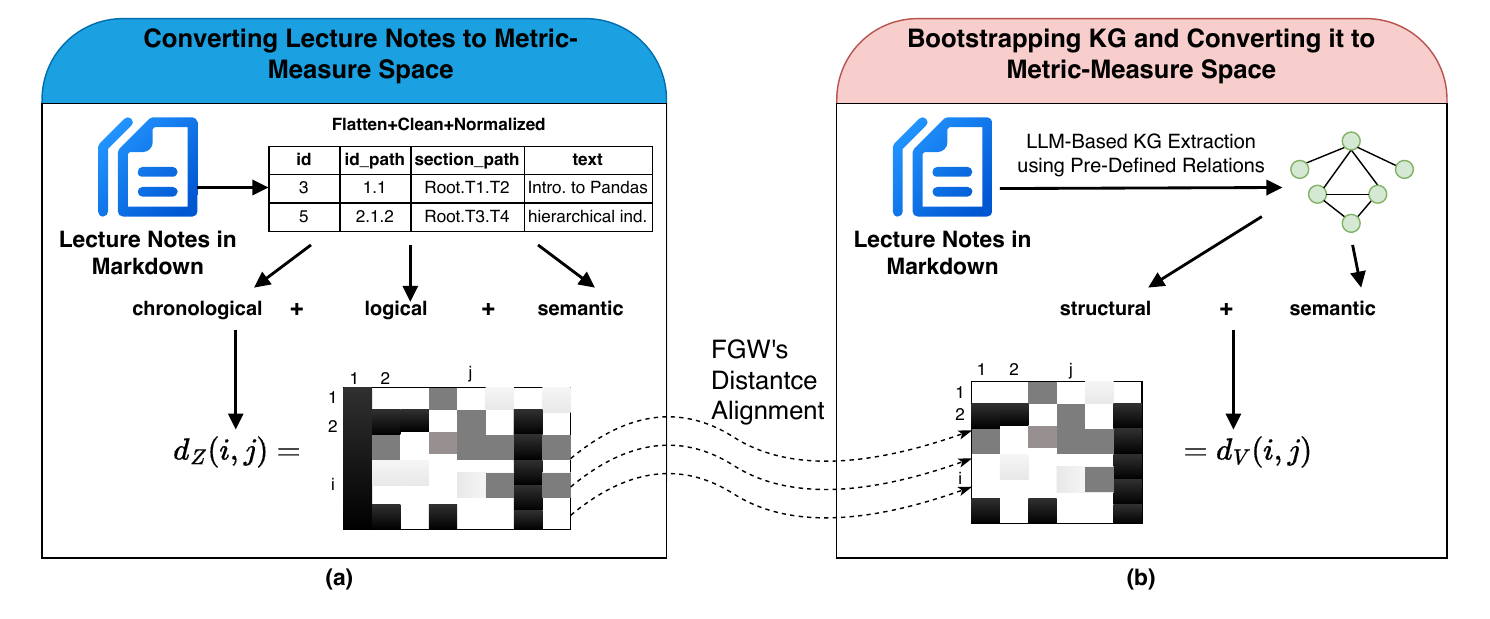} 
        \caption{(a) Lecture notes in markdown are flattened into elementary segments encoding hierarchical context. 
        Text embeddings are computed, and chronological, logical, and semantic distances are combined into a 
        distance matrix $d_Z(i, j)$; (b) An initial knowledge graph (KG) is extracted from the same notes by an LLM under 
        a predefined set of task-oriented relations. Graph structural and node-semantic distances are 
        integrated into a corresponding matrix $d_V(i, j)$.} 
        \label{fig:cold_start} 
\end{figure}

\subsection{Converting Lecture Notes to Metric–Measure Space 
$S = (Z, d_Z, \mu_Z)$ (Figure \ref{fig:cold_start} (a))}
\label{subsec:lecture_mms}

\textbf{Flattening Lecture Notes Structure.}
We assume that lecture notes are described in Markdown format.  
Given markdown description, lecture notes are first parsed into hierarchical JSON trees 
containing fields such as \texttt{id}, \texttt{level}, \texttt{title}, \texttt{content}, and \texttt{children}. 
We recursively traverse this structure to extract \emph{atomic textual units}, 
each corresponding to a minimal self-contained segment of pedagogical meaning. 
Each unit is assigned: 
(1) a chronological index $\mathrm{idx}$ capturing its position within the lecture, 
(2) a hierarchical section path recording the nested heading context, and 
(3) the associated textual content. 
The result is a flattened ordered set $Z = \{ z_i \}_{i=1}^{N}$ representing the lecture corpus.

\textbf{Semantic Embedding.}
Each normalized element $z_i$ is then embedded into a continuous vector space using a pre-trained 
sentence embedding model, producing an embedding matrix 
$E \in \mathbb{R}^{N \times d}$. 
These embeddings encode the local semantic meaning of each lecture segment, 
serving as the foundation for the semantic component of the distance metric $d_Z$.

\textbf{Component Distance Construction.}
For each pair $(z_i, z_j) \in Z \times Z$, we compute three complementary 
distance components that jointly capture temporal, structural, and semantic relations:
\begin{align*}
D_{\text{chron}}(i,j) &= \frac{|\mathrm{idx}_i - \mathrm{idx}_j|}{\max(\mathrm{idx})}, \\
D_{\text{logic}}(i,j) &= 1 - \frac{\mathrm{LCP}(\mathrm{path}_i, \mathrm{path}_j)}{\mathrm{max\_depth}}, \\
D_{\text{sem}}(i,j)   &= \mathrm{clip}\!\bigl(1 - \cos(E_i, E_j), 0, 2\bigr),
\end{align*}
where, $D_{\text{chron}}(i,j)$ encodes narrative separation along lecture order,
$D_{\text{logic}}(i,j)$ captures document hierarchy, with $\mathrm{LCP}$ denoting 
the length of the longest common prefix between two section paths, and 
$D_{\text{sem}}(i,j)$ represents conceptual dissimilarity via cosine similarity between embeddings. 
The clipping function
$clip(x,0,2)=min(max(x,0),2)$ ensures that all distances remain in the valid range.
Each matrix is normalized to $[0,1]$ to ensure comparable scaling across modalities.

\textbf{Combined Lecture Distance.}
The overall distance metric $d_Z$ is constructed as a convex combination of the 
three normalized components:
\begin{equation*}
d_Z(i,j) = 
\alpha_{\text{chron}} D_{\text{chron}}(i,j) +
\alpha_{\text{logic}} D_{\text{logic}}(i,j) +
\alpha_{\text{sem}}  D_{\text{sem}}(i,j),
\end{equation*}
followed by min–max normalization.  
Empirically, we initialize with 
$(\alpha_{\text{chron}}, \alpha_{\text{logic}}, \alpha_{\text{sem}}) = (0.2, 0.3, 0.50)$, 
which balances temporal coherence, document hierarchy, and semantic proximity 
across diverse lecture styles.

\textbf{Measure Definition.}
At cold start, a uniform measure $\mu_Z(z_i) = 1/N$ is assigned over all lecture elements, 
representing equal prior importance. 
Alternative weighting schemes (e.g., length-proportional or section-aware) can be applied 
after refinement, but the uniform initialization provides a neutral and stable baseline 
for rate–distortion optimization.

\subsection{Bootstrapping an Initial Knowledge Graph and Generating
Metric-Measure Space $O = (V, d_V, \mu_V)$ (Figure \ref{fig:cold_start} (b))}
\label{subsec:bootstrapping}

\textbf{Knowledge Graph Extraction (KG).}
We employ an LLM to bootstrap an initial KG from the same lecture notes
with a fixed ontology of \textit{allowed relations}, including 
\texttt{isA}, \texttt{partOf}, \texttt{prerequisiteOf}, \texttt{dependsOn}, \texttt{uses}, 
\texttt{exampleOf}, \texttt{contrastsWith}, \texttt{implies}, \texttt{provedBy}, 
\texttt{produces}, \texttt{consumes}, and \texttt{assessedBy}, among others. 
The prompt instructs the LLM to processes the markdown lecture notes input in three stages: 
(1) segmentation of atomic spans (paragraphs, lists, math, or code blocks), 
(2) identification of salient nodes with canonical labels, definitions, and aliases, and 
(3) extraction of relation-constrained edges grounded in the allowed set.  
Every node and edge is annotated with provenance (section path, line span, and text excerpt) 
and assigned a confidence score in $[0,1]$ with a rationale string, ensuring 
traceability and factual grounding. The result is stored as an undirected 
graph graph $G = (V, E)$, with a node set $V = \{v_i\}_{i=1}^{M}$ 
and an edge set $E \subseteq V \times V$ constrained to the allowed relations. 
 
\textbf{Component Distance Construction.}
We compute two complementary distances are for the knowledge graph metric-measure space:
\begin{flalign*}
D_{\text{struct}}(i,j) &= \mathrm{norm}\bigl(\text{shortest\_path\_length}(v_i,v_j)\bigr), \\
D_{\text{sem}}(i,j) &= \mathrm{clip}\!\bigl(1 - \cos(E_i, E_j), 0, 2\bigr),
\end{flalign*}

where $E_i$ and $E_j$ denote the embedding vectors derived from the node texts 
(label, definition, and up to three aliases).  
The structural distance $D_{\text{struct}}$ reflects topological proximity within the 
graph, shortest path length normalized to $[0,1]$, while the semantic distance 
$D_{\text{sem}}$ quantifies conceptual dissimilarity between node meanings.  
The clipping operator $\mathrm{clip}(x,0,2)$ is the same as before.

\textbf{Combined Knowledge Graph Distance.}
The combined node distance metric is defined as a convex fusion:
\begin{equation*}
d_V(i,j) = 
\gamma_{\text{struct}} D_{\text{struct}}(i,j) + 
\gamma_{\text{sem}}   D_{\text{sem}}(i,j),
\end{equation*}
followed by normalization to $[0,1]$.  
The hyperparameters $(\gamma_{\text{struct}}, \gamma_{\text{sem}})$ 
control the relative importance of relational topology versus semantic similarity; 
a little skewed initialization $(0.4, 0.6)$ is adopted for the cold-start configuration.  
This fused metric encodes both conceptual and relational geometry within the graph.

\textbf{Measure Definition.}
Analogously, a uniform measure $\mu_V(v_i) = 1/M$ 
is assigned across all nodes at initialization, representing equal prior weight.  
Non-uniform weighting (e.g., degree- or centrality-based) can be introduced in later 
iterations when optimizing the rate-distortion objective.

\begin{figure}[h] 
        \centering 
        \includegraphics[width=\columnwidth]{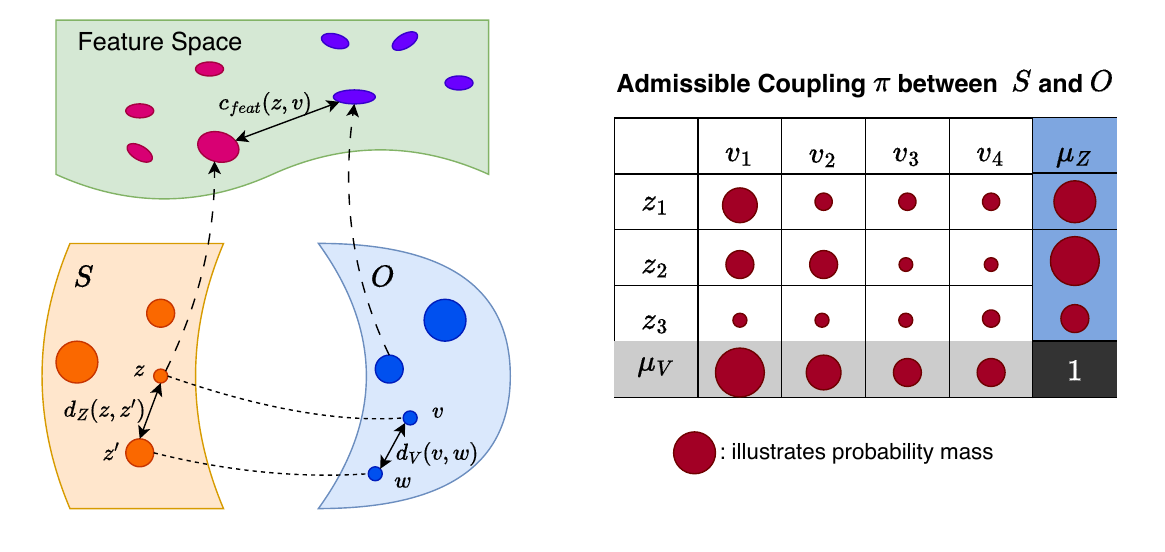} 
        \caption{The FGW alignment jointly minimizes structural and feature mismatches, yielding an optimal coupling $\pi$ 
         between lecture elements and KG nodes. The table on the right visualizes an admissible coupling matrix, 
	where cell intensities represent the transported probability mass. Row and column sums correspond to the marginal 
	distributions $\mu_Z$ and $\mu_V$, respectively.} 
        \label{fig:fgw_coupling} 
\end{figure}

\subsection{Alignment via Fused Gromov-Wasserstein Optimal Transport
(Figure \ref{fig:fgw_coupling})}

Given the source metric space $S = (Z, d_Z, \mu_Z)$ and the knowledge graph space $O = (V, d_V, \mu_V)$, 
along with feature costs, we next compute the fused Gromov-Wasserstein coupling $\pi$ that underlies the 
distortion $D$. Solving the FGW alignment exactly is computationally challenging for large sets, but we 
leverage entropic regularization and Sinkhorn’s algorithm to make it tractable. 

In implementation, we utilized the Python Optimal Transport (POT) \cite{flamary2021pot} 
library’s FGW solver with a 
modest regularization $\epsilon$. This yields a soft alignment: every lecture segment 
is fractionally matched to multiple concept nodes, but typically a few concept 
alignments carry most of the weight for each segment.

The outcome of this alignment step is twofold. First, we get the distortion value 
$D$ for the current knowledge graph. Second, we get 
insights from the coupling $\pi$ about which parts of $Z$ align where 
in $V$ (see Figure \ref{fig:fgw_coupling}). This guides the refinement. 
This alignment step is repeated after each knowledge graph modification 
during refinement. Fortunately, the knowledge 
graph sizes and number of segments in our experiments are not enormous 
(on the order of a few hundreds nodes and segments), the Sinkhorn-based 
FGW computation is quite manageable (a few seconds per iteration). 
If scaling up, one might need more efficient heuristics or minibatch OT approximations \cite{Kalinowski2023}.

\subsection{Refinement Operators in Practice}
\label{subsec:refinements}

For an undirected KG graph $G=(V,E)$, we implement $\mathrm{Rate}(O) \;=\; |V| \;+\; 0.5\,|E|$. 
Given a current FGW coupling $\pi \in\mathbb{R}^{|Z|\times|V|}$ between $S$ and $O$, we iterate local
refinement operations on $O$ while recomputing the KG geometry $d_V$ and the coupling $\pi$ after each edit. 
The refinement operations are implemented as follows.

\textbf{Op-A: Add Under-Represented Concepts and Associated Edges.}
We identify under-covered lecture elements via row-mass of the coupling, $\rho_i=\sum_j \pi_{ij}$, 
and select candidates with $\rho_i<\theta_{\text{add}}$. For each candidate element $z_i$,
we synthesize a new concept $v_{\text{new}}$. We use an LLM to generate an appropriate label 
for $v_{\text{new}}$ and connect it to the rest of the KG nodes 
constrained to the allowed relation set. 
As a fallback, when the LLM cannot propose valid relations, we optionally connect $v_{\text{new}}$ 
to the nearest existing concept by semantic similarity with a low-confidence \texttt{relatedTo} edge.

\textbf{Op-B: Split Overloaded Concepts.}
To detect semantically overloaded nodes, we measure the entropy of each \emph{column} of the coupling,
\[
H_j \;=\; -\sum_{i}\, \tilde{p}_{ij}\,\log \tilde{p}_{ij}
\quad\text{with}\quad 
\tilde{p}_{ij} = \frac{\pi_{ij}}{\sum_{i'} \pi_{i'j}},
\]
and mark $v_j$ for splitting if $H_j>\theta_{\text{split}}$. 
We collect the coupled lecture subset $\{z_i : \pi_{ij}>mean({\pi}_{\cdot j})\}$ 
and run $k$-means ($k=2$) on the lecture embeddings to obtain two groups $A,B$. 
We then create children $v_{j,a}$ and $v_{j,b}$ by cloning $v_j$’s attributes and 
updating definitions with concatenated group texts; names are produced by the same LLM namer, 
with TF-IDF fallback if the LLM returns empty. All edges incident to $v_j$ are 
rewired to both $v_{j,a}$ and $v_{j,b}$, and $v_j$ is removed. 

\textbf{Op-C: Merge Redundant Concepts.}
We scan unordered pairs $(v_i,v_j)\in V$ with high semantic similarity and near-identical coupling profiles. 
Let $E_K$ be the node-embedding matrix and let $\pi_{\cdot i}$ be column $i$ of coupling $\pi$.
We accept a merge if 
\[
\cos(E_{K,i},E_{K,j}) \;\ge\; \theta_{\cos}
\quad\text{and}\quad
\mathrm{KL}_{\mathrm{sym}}(\pi_{\cdot i}\|\pi_{\cdot j}) \;\le\; \theta_{\mathrm{merge}},
\]
where $\mathrm{KL}_{\mathrm{sym}}(p\|q)=\tfrac12(\mathrm{KL}(p\|q)+\mathrm{KL}(q\|p))$. 
We greedily merge $v_j$ into $v_i$, preserving $v_i$’s label/definition and absorbing $v_j$’s label and aliases into $v_i$’s aliases; 
all edges incident to $v_j$ are rewired to $v_i$, and $v_j$ is removed. 
We rebuild $d_V$ and recompute $P$.

\textbf{Op-D: Relationship Updates (Add/Remove).}
\emph{Add:} For each pair of concepts $(v_j,v_k)\in V$, if their top-coupled lecture neighborhoods 
are mutually close in $d_Z$, 
i.e., the mean of $\{d_Z(z_{i_1},z_{i_2}) : z_{i_1}\in \mathrm{Top}_5(j), z_{i_2}\in \mathrm{Top}_5(k)\}$ is below $\theta_{\text{relate}}$, 
we add a \texttt{relatedTo} edge between $(v_j,v_k)$ if absent. 
\emph{Remove:} We prune weakly supported edges by a coupling support score 
$s(a,b)=\bigl(\sum_i \pi_{ia}\bigr)\!\cdot\!\bigl(\sum_i \pi_{ib}\bigr)$; edges with $s(a,b)<\tau$ are removed.

\textbf{Optional LLM Refinements.}
We implement an LLM-based refinement procedure that proposes \emph{new} edges 
\emph{only} from knowledge already present in the KG (labels/definitions/relations), 
with structured rationales and confidence. 

\textbf{Search Procedure and Early Stopping.}
At each iteration we perform a bounded number of local edits (caps on adds/splits/merges), 
recompute $(d_V, \pi)$, and record 
$\mathrm{rate}$, the FGW totals (structure/feature), and the objective ${L}$. 
If ${L}$ improves, we update the incumbent; otherwise, we continue but trigger early stopping when the 
objective change falls below a convergence threshold.

\textbf{Prototype Hyperparameters.}
We use $(\gamma_{\text{struct}},\gamma_{\text{sem}})$ to fuse KG distances; 
$\theta_{\text{add}}$ for under-coverage, $\theta_{\text{split}}$ for entropy-based splitting, 
$\theta_{\cos}$ and $\theta_{\mathrm{merge}}$ for merges, 
$\theta_{\text{relate}}$ for relation additions, and $\tau$ for pruning. 
Sinkhorn parameters $(\lambda_{\text{feat}},\varepsilon, T)$ control the FGW solver. 
All distance components are individually normalized to $[0,1]$ before fusion.

\textbf{Example Scenario (pandas Lecture Segment).}
In one iteration, several lecture elements about hierarchical indexing had low row-mass ($\rho_i<\theta_{\text{add}}$). 
\textbf{Op-A} added a new node \texttt{``MultiIndex basics''} with provenance to the relevant section and an LLM-suggested 
\texttt{uses} edge to \texttt{``stack/unstack''}. 
Subsequently, \textbf{Op-B} flagged \texttt{``indexing''} for split (high column entropy), producing 
\texttt{``set\_index''} and \texttt{``reset\_index''} children with updated definitions; 
edges incident to \texttt{indexing} were duplicated to both new nodes. 
Next, \textbf{Op-C} merged \texttt{``groupby aggregation''} and \texttt{``groupby-agg''} 
(high cosine similarity, low symmetric KL on coupling). 
Finally, \textbf{Op-D} added \texttt{relatedTo(set\_index, sort\_index)} due to proximal lecture neighborhoods and 
pruned a weak \texttt{relatedTo} elsewhere (low coupling support). 
The fused KG distance $d_V$ and coupling $P$ were recomputed, yielding a lower objective $\mathcal{L}$.

\subsection{Evaluation Methods}

Our implementation focuses on evaluating coverage and retrieval practice quality, 
following the definitions in Section \ref{subsec:evaluation_metrics}. 
We measure how refinement affects coverage and compare the quality of 
MCQs generated from the optimized knowledge graph with those derived from the 
raw lecture notes.

%
%
\section{Experiments}
\label{sec:experiments}

\subsection{Dataset and Setup}

We evaluated our approach using lecture materials from 
an introductory Data Science programming course. The dataset consisted of eight weeks of lecture notes originally 
in Jupyter notebooks covering core topics including \texttt{numpy}, \texttt{pandas}, \texttt{data aggregation}, and \texttt{time 
series analysis}. Each notebook was first converted into Markdown format ready for generating a metric-measure space 
described in Section~\ref{subsec:lecture_mms}.
From these lecture notes, we extracted an initial knowledge graph per week using the LLM-based bootstrapping 
method (Section~\ref{subsec:bootstrapping}). 
We then apply the refinement operations as described in Section \ref{subsec:refinements} to generate optimized KG.

\subsection{Key Hyperparameters}
For our dataset, the \textbf{rate term} is defined as 
\[
R(O) = |V| + \frac{|E|}{2},
\]
which typically falls within the range of several dozens to a few hundred, reflecting the structural complexity of the graph. 
In contrast, the \textbf{distortion term} \( D \), defined as the Fused Gromov--Wasserstein (FGW) distance, 
is a small fractional value less than 1. 
To ensure that both terms contribute comparably to the Lagrangian objective
\[
L = R + \beta D,
\]
we set the balancing coefficient \( \beta \) to a large value. 
In our experiments, we use \( \beta = 100 \).

The other hyperparameters are:
\[
\lambda_{\text{feat}} = 0.6, \quad
\theta_{\text{add}} = 0.02, \quad
\theta_{\text{split}} = 0.35, \quad
\theta_{\text{merge}} = 0.12, \quad
\]
\[
\theta_{\text{cos}} = 0.90, \quad
\theta_{\text{relate}} = 0.25, \quad
\tau = 10^{-4}, \quad
\epsilon_{\text{sinkhorn}} = 0.05.
\]

\subsection{Knee Point Identification}

\begin{figure}[h]
    \centering
    \includegraphics[width=0.5\textwidth]{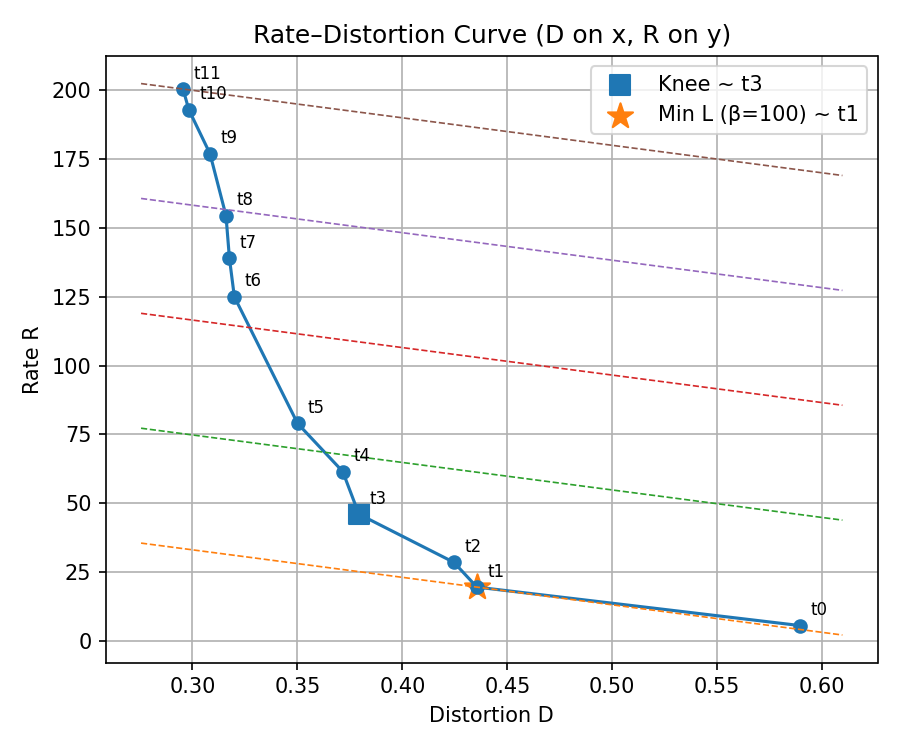}
        \caption{
	Rate--Distortion (RD) curve showing the trade-off between distortion ($D$) and representational 
	rate ($R$) across iterative KG refinements ($t_0$-$t_{11}$). Each blue point denotes a refinement step, 
	with the solid line forming the empirical RD frontier. 
	Dashed lines represent {iso-Lagrangian} contours ($L = R + \beta D$). 
	The blue square marks the {knee point}, the location of maximum curvature corresponding 
	to the most efficient balance between rate and distortion.
}    \label{fig:rd_curve_example}
\end{figure}

To identify the optimal operating point (knee point) along each RD curve, we 
used the  geometric criterion that identifies the refinement step $t_k$ having the maximum perpendicular 
distance to the line segment joining the first and last RD points. The knee 
corresponds to the point of 
diminishing returns where additional refinement yields limited information gain relative to complexity. 
Figure~\ref{fig:rd_curve_example} illustrates a representative RD curve, with the optimized knowledge graph 
(KG) corresponding to the identified knee point.

\subsection{Results and Findings}

\textbf{Coverage Analysis.}
To assess how a knowledge graph (KG) captures the corresponding 
lecture, we computed the \emph{coverage score} which is a number in $[0, 1]$ indicating 
the fraction of lecture segments that are matched to at 
least one KG node under a low feature-distance threshold. Given the matrix of feature distance between
lecture elements and KG nodes, we set the tolerance level to be the 30th 
percentile of the feature-distance distribution. A higher coverage indicates that the KG explains a larger 
portion of the instructional material, while a low value implies conceptual gaps in representation.

\begin{figure}[h]
    \centering
    \includegraphics[width=0.45\textwidth]{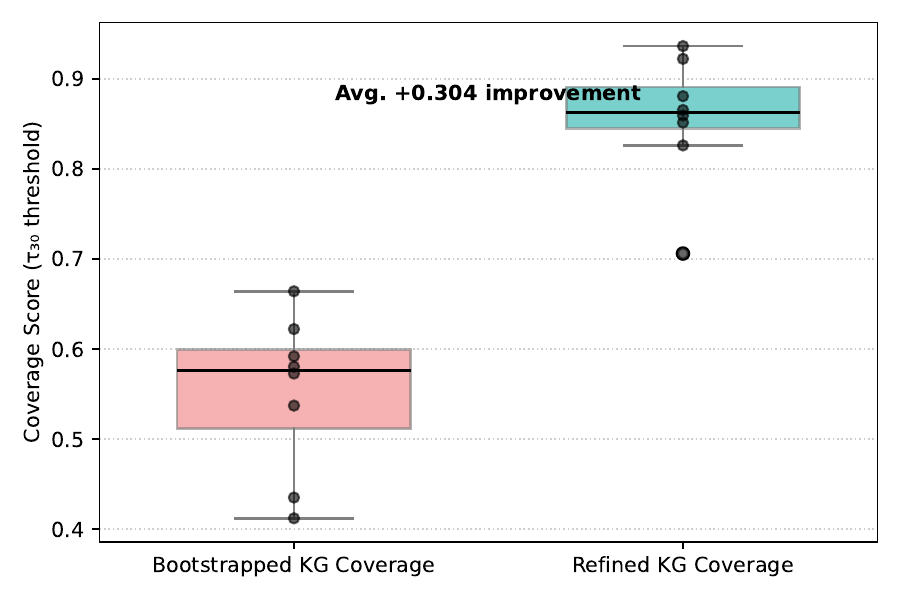}
    \caption{Coverage Improvement from LLM Bootstrapped to RD-Guided Refined Knowledge Graphs.}
    \label{fig:coverage_improvements}
\end{figure}

Figure \ref{fig:coverage_improvements} reports both the coverage of the initial (LLM bootstrapped) 
KGs and the refined KGs selected at the RD knee point. Across all lectures, coverage increased substantially after rate-distortion 
guided refinement, confirming that the iterative FGW coupling and optimization enhanced 
semantic alignment with the lecture materials. On average, the refined KGs achieved a $\mathbf{+0.304}$  absolute improvement in 
coverage, significantly increasing the explained fraction of lecture content.

\textbf{Rate-Distortion Behavior.}
Across lectures, the RD curve consistently exhibited a steep initial decline in distortion 
(high information gain per added complexity) followed by a gradual flattening as refinements saturated. 
This result validates the theoretical expectation of diminishing returns in semantic 
enrichment beyond the knee (see an example in Figure \ref{fig:rd_curve_example}). The knee point thus represents an optimal 
compromise between representational compactness and semantic fidelity.
On average,  a knee point corresponds to 
a $\approx$50\% reduction in distortion with only $\approx$30\% of the total 
rate cost. Beyond this point, additional refinements marginally improved coverage 
but incurred disproportionate complexity growth.

\textbf{Comparison of KG-Based and Lecture-Based MCQ Quality.}
We evaluated both the KG-based and lecture-based multiple-choice question (MCQ) sets 
for each lecture using a set of 15 criteria covering \emph{factual accuracy}, \emph{content alignment}, \emph{clarity}, 
\emph{distractor quality}, etc. Each criterion was scored on a 5-point Likert scale, 
where higher scores indicate better quality with respect to that criterion. 
The full set of criteria is summarized in Table~\ref{tab:criterion_improvement}.

For each lecture, we prompted the LLM to generate 25 MCQs directly from the raw 
lecture notes and another 25 MCQs from the refined KG. Each MCQ was then independently 
evaluated against all 15 criteria. For example, the criterion \texttt{no\_hallucinations} was 
defined as: \emph{``Content is grounded in the source material, not invented."} It was evaluated as:
``\texttt{1}: \emph{mostly invented or irrelevant to lecture materials}",
``\texttt{2}: \emph{several invented or off-topic statements}",
``\texttt{3}: \emph{some content not grounded in lecture materials}",
``\texttt{4}: \emph{mostly grounded with only minor unsupported details}",
``\texttt{5}: \emph{fully grounded in the source material with no hallucination.}"

In total, 400 MCQs were evaluated (8 lectures × 50 questions per lecture) 
using the 1-5 scoring scale. The aggregated results are presented in Table~\ref{tab:kg_mcq_results}. 
Across all lectures, KG-based MCQs consistently outperformed those generated directly 
from lecture notes. On average, KG-based generation achieved a total quality 
score improvement of approximately $\mathbf{+15\%}$.

\begin{table}[h]
\centering
\caption{Average MCQ evaluation results (15-criterion rubric).}
\label{tab:kg_mcq_results}
\begin{tabular}{lcc}
\toprule
Lecture \#& KG-based Avg. Score & Lecture-based Avg. Score \\
\midrule
1 & 4.68 & 4.17\\
2 & 4.78 & 4.25 \\
3 & 4.76 & 4.05 \\
4 & 4.67 & 4.12 \\
5 & 4.63 & 3.92 \\
6 & 4.81 & 4.09 \\
7 & 4.65 & 3.99 \\
8 & 4.66 & 4.12 \\
\midrule
\textbf{Mean} & \textbf{4.70} & \textbf{4.08} \\
\bottomrule
\end{tabular}
\end{table}

\textbf{Criterion-Level Improvements.}
Table~\ref{tab:criterion_improvement} reports average improvements per evaluation criterion. 
The largest gains occurred in \texttt{no\_answer\_hints} (+1.805), \texttt{hard\_to\_guess} (+1.210), 
\texttt{no\_duplicates} (+1.190), and \texttt{option\_balance} (+1.110). 
These metrics directly capture the benefits of refined, semantically structured 
KGs. Interestingly, the KG-based MCQs are worse compared to the MCQs generated from raw lecture notes  
on the criteria \texttt{incorrect\_assumption} (-0.010), 
\texttt{no\_hallucinations} (-0.150), and
\texttt{unambiguous\_options} (-0.185).
One plausible explanation is that the inherently concise and abstract nature of the knowledge graph 
offers limited textual cues for LLM to leverage during MCQ generation. 
In future research, we will investigate strategies to enhance the semantic richness and 
contextual grounding of KGs to better support high-quality, unambiguous, and factually 
consistent MCQ generation.

\begin{table}[h]
\centering
\caption{The 15 criteria and per-criterion average improvement ($\Delta = \text{KG} - \text{Lecture}$) across all lectures.}
\label{tab:criterion_improvement}
\begin{tabular}{llc}
\toprule
Criterion \# & Criterion & Avg. $\Delta$ Score \\
           &  Name & (1-5 scale) \\
\midrule
1 & no\_answer\_hints      &      1.805 \\
2 & hard\_to\_guess          &    1.210 \\
3 & no\_duplicates          &    1.190 \\
4 & option\_balance        &     1.110 \\
5 & plausible\_distractors  &    1.095 \\
6 & cognitive\_engagement  &     0.995 \\
7 & grammar\_fluency          &  0.925 \\
8 & discrimination\_ability   &  0.590 \\
9 & unambiguous\_question  &     0.370 \\
10 & correct\_key\_answer       &  0.190 \\
11 & unambiguous\_correctness  &  0.090 \\
12 & factual\_accuracy         &  0.085 \\
13 & incorrect\_assumption   &   -0.010 \\
14 & no\_hallucinations       &  -0.150 \\
15 & unambiguous\_options  &     -0.185 \\
\midrule
 & \textbf{Average Improvement} & \textbf{+0.62} \\
\bottomrule
\end{tabular}
\end{table}

\subsection{Summary}

The experiments demonstrate that rate-distortion guided knowledge graph (KG) 
refinement identifies a stable trade-off point (knee) that balances richer 
representation with less information loss. The refined KGs produced at this point 
lead to improved question quality. 
These findings support rate-distortion analysis as both a theoretical foundation 
and a practical approach for controlling 
LLM-based knowledge graph refinement in educational contexts.

%
%
\section{Conclusion}
\label{sec:conclusion}

We presented a novel framework for constructing and refining knowledge graphs from lecture notes, 
grounded in the principles of rate-distortion theory and optimal transport. 
By formalizing the problem as one of compressing lecture content into an knowledge graph, 
we introduced a quantifiable trade-off. Every additional concept or link (increasing complexity) 
should yield a proportional reduction in distortion (information loss), and vice versa.

Using fused Gromov-Wasserstein optimal transport, we can measure the distortion 
between the lecture material and the knowledge graph in a way that accounts for both semantic content 
and relational structure. This presents a powerful feedback signal to drive knowledge graph refinement. Our 
prototype on Data Science lectures demonstrated that this signal can successfully guide the addition 
of missing concepts, the merging of duplicates, the pruning of duplicated nodes, and the adjustment of 
relations, resulting in a much-improved knowledge graph. The final knowledge graph achieved a good 
balance. It was compact yet faithful to 
the lecture content.

This work contributes a new perspective to both the knowledge graph learning literature and AI in education. 
Rather than relying purely on heuristics or the output of a large language model, our method provides a 
principled objective and produces interpretable, data-backed recommendations for knowledge graph edits. 
It effectively bridges information theory and knowledge graph engineering, showing that concepts from 
rate-distortion can be meaningfully applied outside of traditional compression domains, in this case to 
compress knowledge. The success of our initial experiment lays a foundation for more extensive studies 
and applications. We believe that such an approach can greatly assist in creating high-quality knowledge 
graphs, which in turn can improve AI-powered learning systems, personalized curricula, 
and student learning outcomes.

%
%
\appendix

The source notebooks and data used in the study are 
accessible in the public GitHub repository: 
https://github.com/ailadudragon/RD\_FGW\_KG\_Refinement .

%
%
\bibliographystyle{unsrt}

\end{document}